\documentclass[12pt, a4paper, reqno]{amsart}

\usepackage{amssymb}
\usepackage[english]{babel}
\usepackage{blindtext}
\usepackage[tableposition=top]{caption}
\usepackage{comment}
\usepackage{csquotes}
\usepackage[inline]{enumitem}
\usepackage{float}
\usepackage[T1]{fontenc}
\usepackage{mathtools}
\usepackage{physics}
\usepackage{tikz-cd}

\usepackage{hyperref}
\usepackage[capitalize]{cleveref}
\hypersetup{
    colorlinks=true,
    allcolors=black,
}

\setlist[enumerate]{topsep=5pt,itemsep=-0.5ex,partopsep=0ex,parsep=1ex}

\theoremstyle{plain}

\newtheorem{theorem}{Theorem}[section]

\theoremstyle{definition}

\theoremstyle{remark}
\newtheorem{remark}[theorem]{Remark}

\title{Flow matching on homogeneous spaces}

\author{Francesco Ruscelli}
\address{Mathematical Institute, University of Heidelberg, 69120 Heidelberg, Germany}
\email{fruscelli@mathi.uni-heidelberg.de}

\begin{document}

\begin{abstract}
    We propose a general framework to extend Flow Matching to homogeneous spaces, i.e.\ quotients of Lie groups. Our approach reformulates the problem as a flow matching task on the underlying Lie group by lifting the data distributions. This strategy avoids the potentially complicated geometry of homogeneous spaces by working directly on Lie groups, which in turn enables us reduce the problem to a Euclidean flow matching task on Lie algebras. In contrast to Riemannian Flow Matching, our method eliminates the need to define and compute premetrics or geodesics, resulting in a simpler, faster, and fully intrinsic framework.
\end{abstract}

\maketitle

\section{Introduction}
The aim of generative models is to learn the probability distribution of a given dataset. One way to achieve this is to take a noise distribution and somehow change into the desired distribution. A geometric framework for doing this, called Continuous Normalizing Flows, was proposed by Chen et al.\ in 2018 \cite{chen2018}. The idea of CNFs is to learn a vector field whose flow pushes the noise distribution forward to the target distribution (i.e.\ the one corresponding to the dataset). In other words, if \( \mu_0 \) is the probability measure corresponding to the noise and \( \mu_1 \) is the probability measure of the dataset, CNFs learn a vector field \( v_t \), with \( t \in [0, 1], \) whose flow \( \varphi_t \) has the property
\begin{equation*}
    \varphi_{t*} \mu_0 = \mu_1.
\end{equation*}

Although CNFs have been very successful, they can be computationally expensive because they require numerical simulation of ODEs during both training and inference. A much more scalable framework was proposed by Lipman et al.\ \cite{lipman2023} and goes by the name of Flow Matching. It builds on CNFs by retaining the overall structure of the model but it modifies the training objective by making it more tractable. Crucially, FM does not require any simulations during training. More specifically, suppose we want to train a neural network \( u^{\theta} \) and that \( X_0, X_1 \) are random variables distributed according the noise and the dataset distributions respectively. Denote by \( v_t \) the vector field we are interested in learning, \( \varphi_t \) its flow and let \( X_t = \varphi_t(X_0) \). The naive FM loss is given by
\begin{equation*}
    \mathcal{L}_{\mathrm{FM}}(\theta) = \mathbb{E} \big[ || u_t^\theta(X_t) - v_t(X_t) ||^2 \big],
\end{equation*}
where the expectation is taken with respect to both \( X_t \) (whose distribution depends on that of \( X_0 \)) and \( t \sim U(0, 1) \).
CNFs do not use this loss function since we usually do not have access to the true vector field \( v_t \). Instead, they optimize the neural network based on \( \varphi_1 \), i.e.\ the flow at time \( t = 1 \), which requires simulating the ODE during training. Flow matching bypasses this computational hurdle by conditioning \( v_t \) on the endpoint (or on some other random variable if one desires a more general framework, as in \cite{tong2024}). The loss function then takes the form
\begin{equation*}\label{eq:euclidean_loss}
    \mathcal{L}_{\mathrm{CFM}}(\theta) = \mathbb{E} \big[ || u_t^\theta(X_t) - v_t(X_t | X_1) ||^2 \big].
\end{equation*}

For instance, if the dataset lies in some Euclidean space \( \mathbb{R}^d \), we can interpolate between points in the noise distribution and dataset points by line segments. An easy computation (see \cite[Theorem 3]{lipman2023}) shows that the vector field generating such a flow is given by
\begin{equation*}
    v_t(x | x_1) = \frac{x - x_1}{1-t}.
\end{equation*}
In other words, integrating \( v_t(x|x_1) \) will take \( x \) to \( x_1 \).

\medskip

From the inception of CNFs \cite{chen2018}, several ideas aimed at improving their training effiency have been put forth (using augmentation \cite{dupont2019}, regularization techniques \cite{yang2020}, \cite{finlay2020}, \cite{onken2021}, \cite{tong2020}, \cite{kelly2020} and stochastic sampling of integration times \cite{du2022}. See also \cite{rozen2021}, \cite{ben-hamu2022}, \cite{grathwohl2019}), none of which however as successful as flow matching \cite{lipman2023}.
Originally developed by Lipman et al.\ on Euclidean space, flow matching has received a lot of attention recently and several improvements and generalizations have been proposed. Chen and Lipman \cite{chen2024} introduced Riemannian flow matching to deal with data lying on general Riemannian manifolds. Tong et al.\ \cite{tong2024} suggested using optimal transport in order to find the best probability paths for the model to learn. A different approach based on variational inference was proposed by Eijkelboom et al.\ \cite{eijkelboom2024} and later extended by Zaghen et al.\ \cite{zaghen2025} to the Riemannian setting. Refinements and applications of variational flow matching were explored in \cite{guzman2025} and \cite{eijkelboom2025}. More recently, flow matching on Lie groups \cite{sherry2026} was introduced as a fully intrinsic, simpler replacement for Riemannian flow matching when working on Lie groups. Finally, \cite{klein2023} studied how to incorporate equivariance into flow matching when dealing with data displaying certain symmetries.

\medskip

The purpose of this work\footnote{The code is available at \url{https://github.com/fresh999/HomogeneousFM}.} is twofold. First, we revisit flow matching on Lie groups \cite{sherry2026} and we rephrase it in a different fashion, namely as flow matching on Lie algebras. We feel our formulation is somewhat conceptually cleaner and allows for a smoother implementation. As in \cite{sherry2026}, the discussion is limited to Lie groups whose exponential map is surjective (for instance compact or nilpotent groups) or to situations in which the datasets are known to lie in the image of the exponential map. Second, we describe a general way of performing flow matching on manifolds that can be seen as homogeneous spaces, namely spaces of the form \( G/H \), where \( G \) is a Lie group and \( H \) is a Lie subgroup. The general idea is to lift the problem of flow matching on \( G/H \) to one on \( G \) and then project the results back down to the quotient. This allows us to avoid the (potentially complicated) geometry of \( G/H \) by going to the much \enquote{flatter} space \( G \). Since most Riemannian manifolds do not have simple closed-form geodesics, our framework also allows us to avoid the computational hassle of computing geodesics. Finally, the algebraic structure of \( G \) can also be used to reduce the model's complexity.

As a proof of concept, we test our framework in two cases:
\begin{enumerate}
    \item \( \operatorname{SL}(2, \mathbb{R}) / \operatorname{SO}(2, \mathbb{R}) \), which is diffeomorphic to the Siegel upper half-plane \( \mathbb{H} \);
    \item \( \operatorname{SO}(3, \mathbb{R}) / \operatorname{SO}(2, \mathbb{R}) \), which is diffeomorphic to the sphere \( S^2 \).
\end{enumerate}

We remark that the former space has already been studied in the context of information geometry, for instance in \cite{barbaresco2013}, \cite{nielsen2020} and in \cite{tumpach2026} for the infinite-dimensional case.

\subsection*{Acknowledgements}
I would like to thank the organizers of the \( 2025 \) GSI (\enquote{Geometric Science of Information}) conference, where this work was originally conceived, for providing an inspiring platform to learn and share ideas. I would also like to thank Alice Barbora Tumpach for fruitful exchanges.
Finally, I am indebted to Rita Fioresi and Ferdinando Zanchetta for introducing me to the field of geometric deep learning, as well as for their guidance and feedback in preparing this work.

\section{Flow matching on Lie groups and homogeneous spaces}
\label{sec:hom_fm}
In this section we revisit the general setting of flow matching on Lie groups as in \cite{sherry2026}. We then use our framework to tackle the question of performing flow matching on homogeneous spaces.
We refer the reader to \cite{varadarajan1984} for a complete account of Lie theory.

\subsection*{Lie groups}
Let \( G \) be a Lie group and consider two probability measures \( \mu_0, \mu_1 \) on \( G \). The goal of flow matching on \( G \) is learning a vector field \( X_t \colon [0, 1] \times G \to TG \) whose flow \( \varphi_t \) pushes \( \mu_0 \) forward to \( \mu_1 \). In order to achieve this using a neural network \( u^\theta \), \cite{sherry2026} proposes the loss function
\begin{equation}\label{eq:lie_loss}
    \mathcal{L}_{\mathrm{CFM}}(\theta) = \mathbb{E} \big[ || u^\theta(g_t) - X_t(g_t | g_1) ||^2 \big],
\end{equation}
where \( t \sim U(0, 1) \), \( g_t = g_0 \exp \big( \log (g_0^{-1} g_1) \big) \) and \( g_0 \) is sampled according to \( \mu_0 \). This loss directly generalizes the Euclidean conditional flow matching loss by replacing line segments by exponential curves. We also note that the norm appearing in \eqref{eq:lie_loss} can a priori be arbitrary but, as noted in \cite{sherry2026}, a geometrically natural choice is to choose a left-invariant one, that is a metric obtained by left-translating a scalar product on the Lie algebra of \( G \).

\medskip

Let us now reformulate this purely in terms of Euclidean flow matching. Whenever we sample elements \( g_0, g_1 \) from \( \mu_0, \mu_1 \) respectively, our assumptions on the surjectivity of \( \exp \) (or the data points lying in the range of \( \exp \)) allow us to pick elements \( x_0, x_1 \in \mathfrak{g} = \operatorname{Lie}(G) \) such that \( g_0 = \exp(x_0) \) and \( g_1 = \exp(x_1) \). In other words, the probability distributions on \( G \) induce probability distributions on \( \mathfrak{g} \). Hence, the whole problem of interpolating between \( g_0 \) and \( g_1 \) in a geometrically sensible way reduces to interpolating between \( X_0 \) and \( X_1 \) in \( \mathfrak{g} \), which is much easier given that \( \mathfrak{g} \) is a vector space.

\begin{remark}
    If we choose a left-invariant metric on \( G \), it is easy to see that performing Euclidean flow matching on \( \mathfrak{g} \) is equivalent to performing flow matching on \( G \) using the loss function \eqref{eq:lie_loss}. Still, we think it is helpful to frame the discussion in terms of flow matching on \( \mathfrak{g} \) because it makes the implementation smoother.  For instance, it is typically easier to encode elements of the Lie algebra using \( \dim \mathfrak{g} \) parameters than to encode elements of the Lie group directly.
\end{remark}

We can thus perform Euclidean flow matching on \( \mathfrak{g} \) by defining \( x_t = (1-t) x_0 + t x_1 \) and the conditional vector field
\begin{equation*}
    v_t(x | x_1) = \frac{x - x_1}{1-t}
\end{equation*}
and regressing our neural network against \( v_t \) using the loss \eqref{eq:euclidean_loss}.
The desired results on \( G \) can finally be obtained by simulating the ODE defined by trained model on \( \mathfrak{g} \), extracting the flow at time \( t = 1 \) and then exponentiating back to \( G \).

\subsection*{Homogeneous spaces}
Let now \( M \) be a manifold that can be written as a homogeneous space \( G/H \) and suppose we have two distributions on \( M \). In order to perform flow matching on \( M \) we lift the distributions to \( G \) and flow match them on \( G \). The desired results are then obtained by projecting back to \( M \).
Let us illustrate this with two examples:
\begin{enumerate}
    \item \( \operatorname{SL}(2, \mathbb{R}) / \operatorname{SO}(2, \mathbb{R}) \cong \mathbb{H} \), where \( \mathbb{H} \) is the Siegel upper half-plane. The diffeomorphism can be seen as follows. \( \operatorname{SL}(2, \mathbb{R}) \) acts on \( \mathbb{H} \) by M\"obius transformations:
        \begin{equation*}
            \begin{pmatrix}
                a & b \\
                c & d
            \end{pmatrix}
            z = \frac{az + b}{cz + d},
        \end{equation*}
        where we view the upper half-plane as the set of complex numbers \( z \) with positive imaginary part. It is easy to see that this action is transitive and that the stabilizer of \( i \) is \( \operatorname{SO}(2, \mathbb{R}) \subseteq \operatorname{SL}(2, \mathbb{R}) \). Quotienting out by the stabilizer gives the desired diffeomorphism.

        The bundle
        \begin{equation*}
            \operatorname{SL}(2, \mathbb{R}) \to \operatorname{SL}(2, \mathbb{R}) / \operatorname{SO}(2, \mathbb{R}) \cong \mathbb{H},
        \end{equation*}
        is trivial, a global section being
        \begin{equation*}
            \begin{aligned}
                &\mathbb{H} \to \operatorname{SL}(2, \mathbb{R}) \\
                &(x, y) \mapsto
                \begin{pmatrix}
                    y^{1/2} & x y^{-1/2} \\
                    0        & y^{-1/2}
                \end{pmatrix}.
            \end{aligned}
        \end{equation*}
        If we have distributions on \( \mathbb{H} \), we can use this section to map them to \( \operatorname{SL}(2, \mathbb{R}) \), flow match them and then project back.

    \item \( \operatorname{SO}(3, \mathbb{R}) / \operatorname{SO}(2, \mathbb{R}) \cong S^2 \). In this case the diffeomorphism can be seen by noting that \( \operatorname{SO}(3, \mathbb{R}) \) acts transitively on \( S^2 \) by left multiplication and that the stabilizer of \( e_3 = (0, 0, 1) \) is given by
        \begin{equation*}
            \left\{
                \begin{pmatrix}
                    A & 0 \\
                    0 & 1
                \end{pmatrix}
                \colon A \in \operatorname{SO}(2, \mathbb{R})
            \right\} \cong \operatorname{SO}(2, \mathbb{R}).
        \end{equation*}

        In this case the bundle \( \operatorname{SO}(3, \mathbb{R}) \to \operatorname{SO}(3, \mathbb{R}) / \operatorname{SO}(2, \mathbb{R}) \cong S^2 \) is not trivial, but we can make it trivial by removing the north pole from \( S^2 \) (in practice, this is not a problem). A section is then given by
        \begin{equation*}
            \begin{aligned}
                \mathbb{R}^2 &\cong S^2 \setminus \{ e_3 \} \to \operatorname{SO}(3, \mathbb{R}) \\
                & x \mapsto I + (\sin{\theta}) K + (1 - \cos{\theta}) K^2,
            \end{aligned}
        \end{equation*}
        where \( K \) is the matrix corresponding to the cross product with the axis \( k = e_3 \times x \) and \( \theta = \arccos(\langle e_3, x \rangle) \) (this is known as Rodrigues' rotation formula).
\end{enumerate}

\begin{remark}
    In practical applications the data could either be given as elements of the quotient space \( G/H \), in which case we have to find a section as above, or as representatives in \( G \). In the latter case, we can directly perform Lie group flow matching.
\end{remark}

In the two toy examples we presented there is no need to perform flow matching following our strategy. Indeed, \( \mathbb{H} \) is convex and hence Euclidean flow matching can be directly applied, while \( S^2 \) is a simple enough Riemannian manifold that Riemannian flow matching can be carried out without trouble. However, the geometry of \( G/H \) could in general be very complicated, in which case lifting the problem to \( G \) allows us to completely sidestep any geometric and computational hurdles (such as computing premetrics and geodesics as in \cite{chen2024}), since we have shown that flow matching on a Lie group is basically equivalent to flow matching on its Lie algebra.

\section{Experiments}
\label{sec:experiments}
In this section we show the results of our experiments on \( \operatorname{SL}(2, \mathbb{R}) / \operatorname{SO}(2, \mathbb{R}) \) and \( \operatorname{SO}(3, \mathbb{R}) / \operatorname{SO}(2, \mathbb{R}) \). Similarly to \cite{lipman2023}, we took checkerboard distributions on \( \mathbb{H} \) and \( S^2 \) and mapped them to \( \operatorname{SL}(2, \mathbb{R}) \) and \( \operatorname{SO}(3, \mathbb{R}) \), as explained in \cref{sec:hom_fm}. We then performed Lie group flow matching and projected the results back to \( \mathbb{H} \) and \( S^2 \). In order to learn the vector field transforming the noise into the target distribution, we used an MLP with \( 5 \) layers of width \( 512 \). Inference was performed by numerically integrating the predicted vector field using the midpoint scheme.

For each homogeneous space we carried out three experiments:
\begin{enumerate}
    \item flow matching in a Lie-agnostic way. Consider for instance the case of \( \operatorname{SL}(2, \mathbb{R}) \). We lifted the checkerboard distribution and then performed Euclidean flow matching on \( \mathbb{R}^4 \supseteq \operatorname{SL}(2, \mathbb{R}) \). This ignores the fact that the lifted distribution lies on \( \operatorname{SL}(2, \mathbb{R}) \) and treats it as a distribution on \( \mathbb{R}^4 \) supported on a subset of measure zero;
    \item \label{itm:ii} Lie group flow matching as explained in \cref{sec:hom_fm}, that is we took the induced distributions on the Lie algebras \( \mathfrak{sl}_2 \) and \( \mathfrak{so}_3 \) and performed Euclidean flow matching on them;
    \item similar to the previous experiment, but we first encoded the Lie algebra elements in a space of parameters (as many as the dimension of the Lie algebra) and performed flow matching on the parameters. This reduces the model's complexity. In the case of \( \operatorname{SO}(3, \mathbb{R}) \) for instance, \( \dim \mathfrak{so}_3 = 3 \). While in experiment \ref{itm:ii} we stored elements of \( \mathfrak{so}_3 \) as \( 3 \times 3 \) matrices, encoding allows us to store them as elements of \( \mathbb{R}^3 \).
\end{enumerate}

The following are the results for \( \operatorname{SL}(2, \mathbb{R}) / \operatorname{SO}(2, \mathbb{R}) \) and \( \operatorname{SO}(3, \mathbb{R}) / \operatorname{SO}(2, \mathbb{R}) \). Note that since the exponential map of \( \operatorname{SL}(2, \mathbb{R}) \) is not surjective, we generated the noise and checkerboard distributions in such a way as to ensure that they lie in the image of the exponential map. We also stereographically projected the distributions on \( S^2 \) to the plane for ease of visualization.

\begin{figure}[H]
    \centering
    \includegraphics[scale=0.29]{"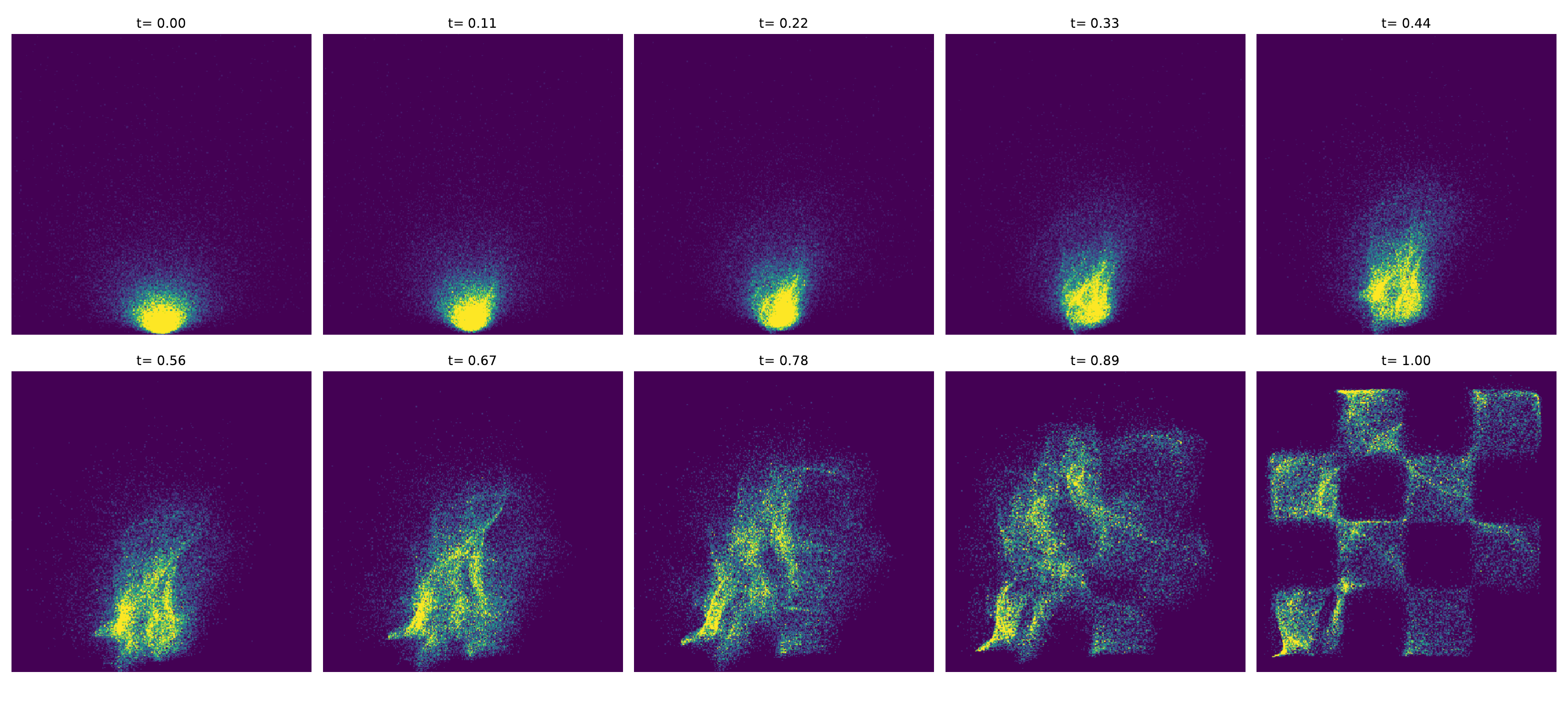"}
    \includegraphics[scale=0.29]{"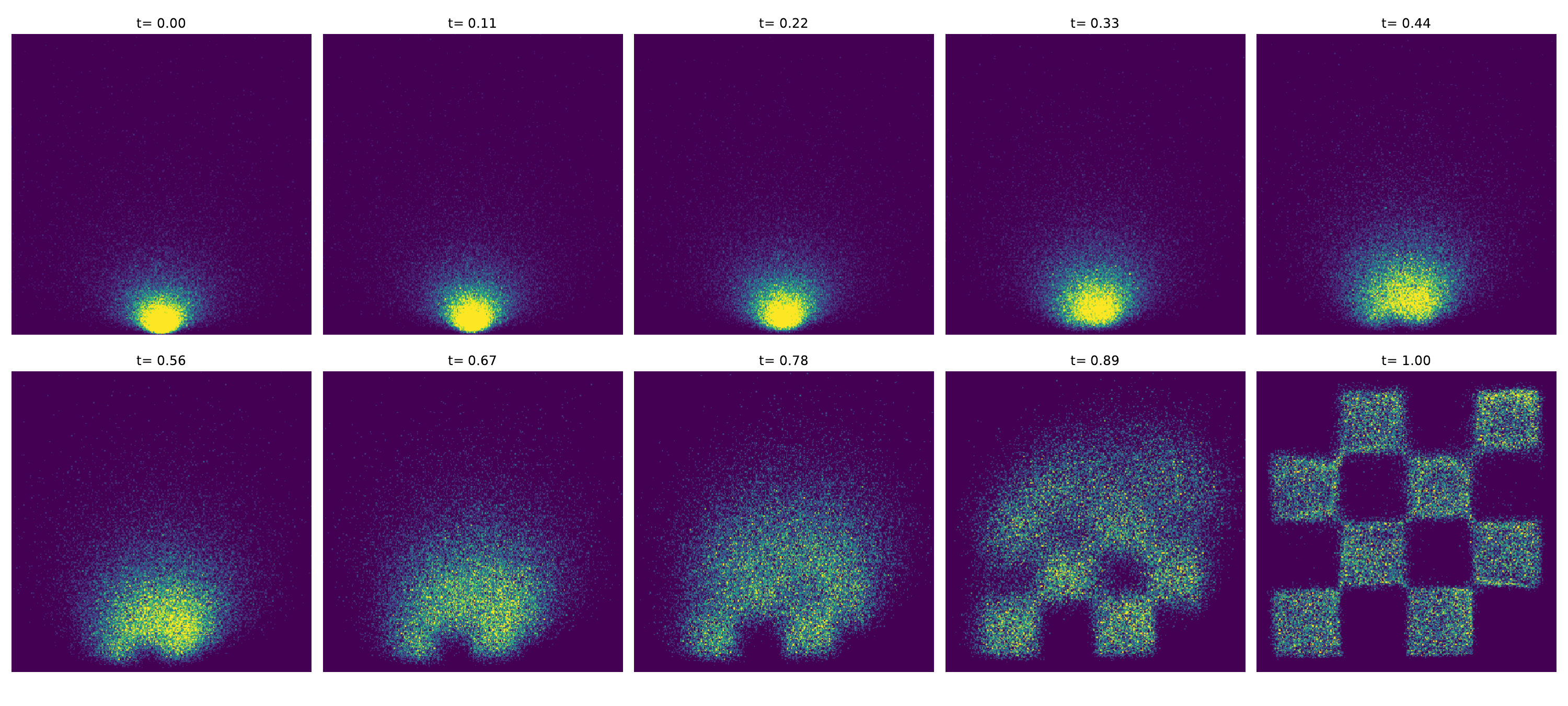"}
    \includegraphics[scale=0.29]{"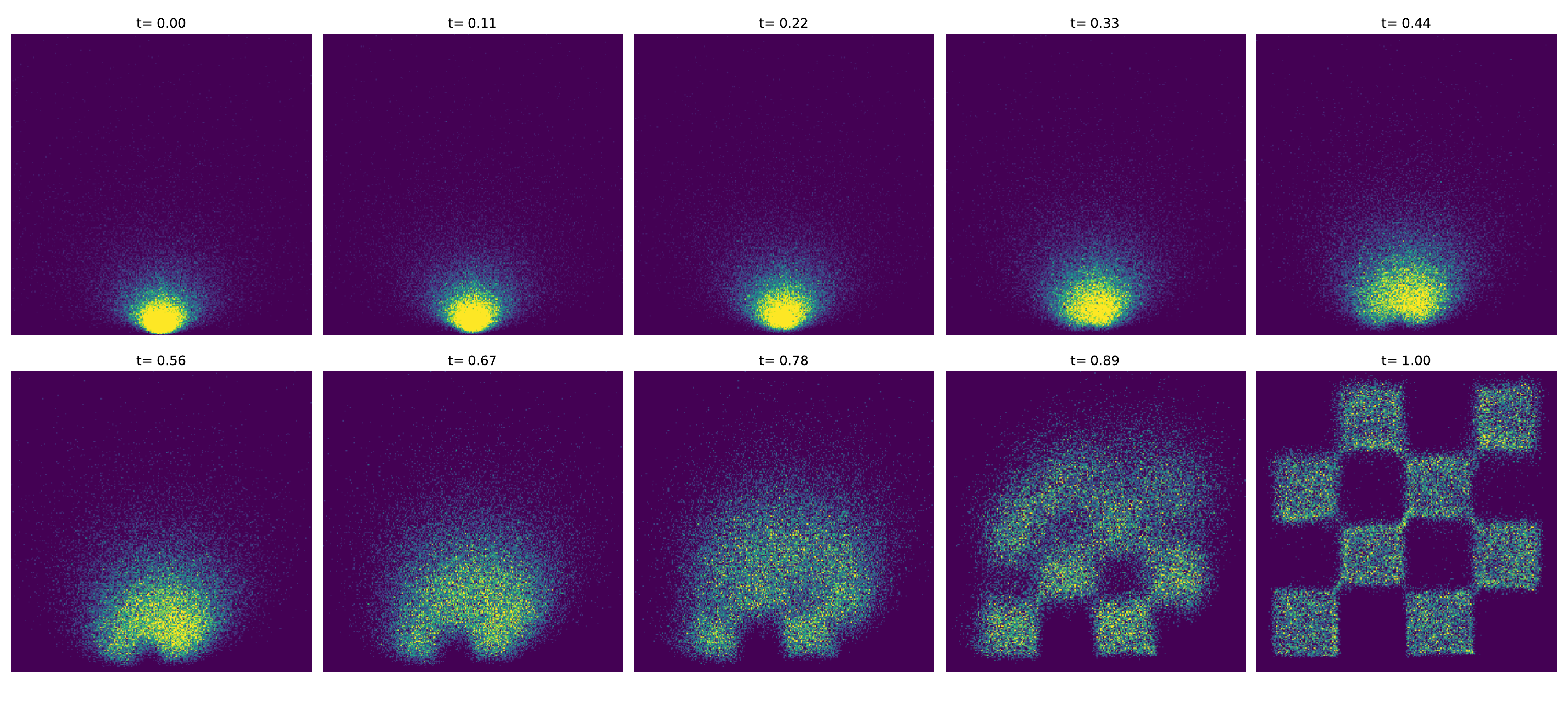"}
    \caption{Flow matching on \( \operatorname{SL}(2, \mathbb{R}) / \operatorname{SO}(2, \mathbb{R}) \cong \mathbb{H} \). Top: Lie-agnostic model. Middle: Lie group model. Bottom: Lie group model with parameter encoding.}
\end{figure}

\begin{figure}[H]
    \centering
    \includegraphics[scale=0.29]{"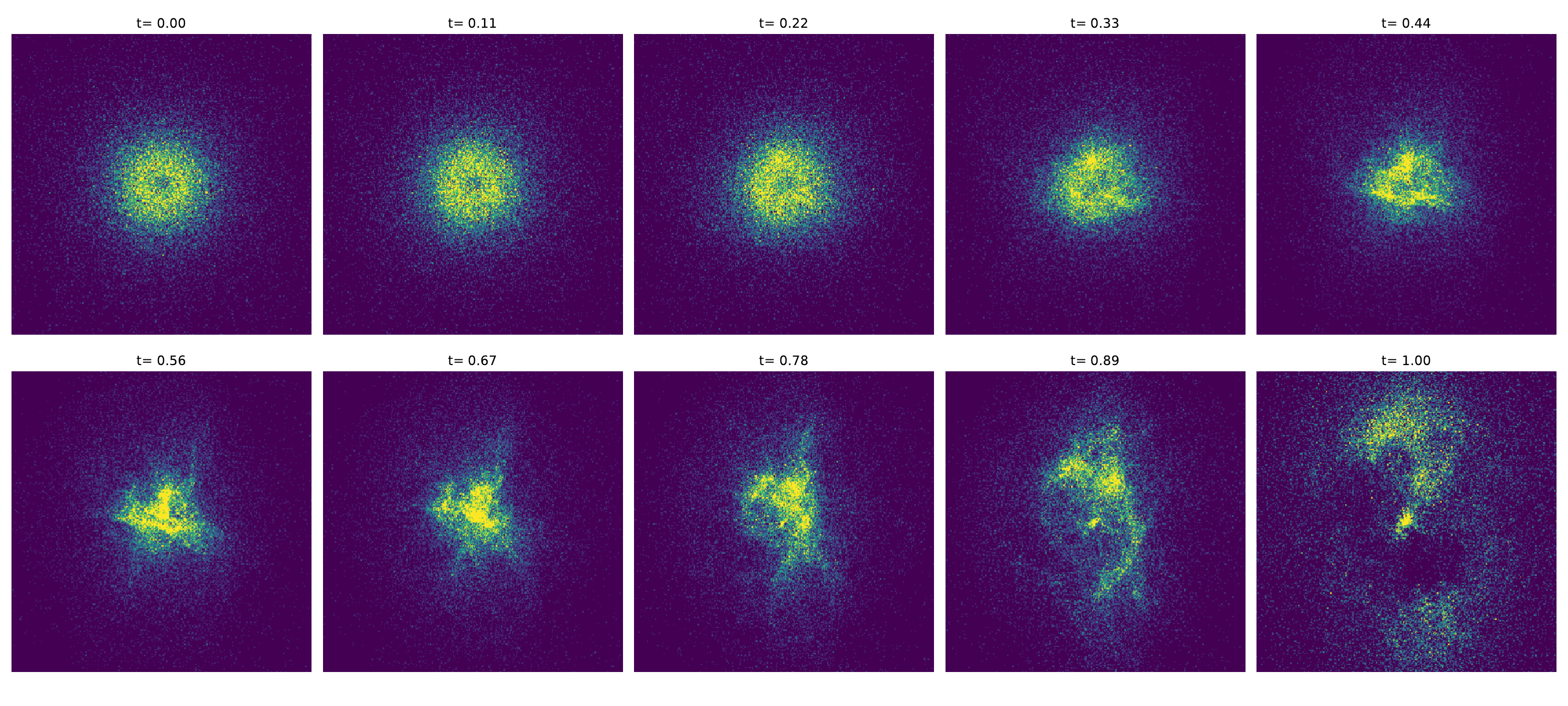"}
    \includegraphics[scale=0.29]{"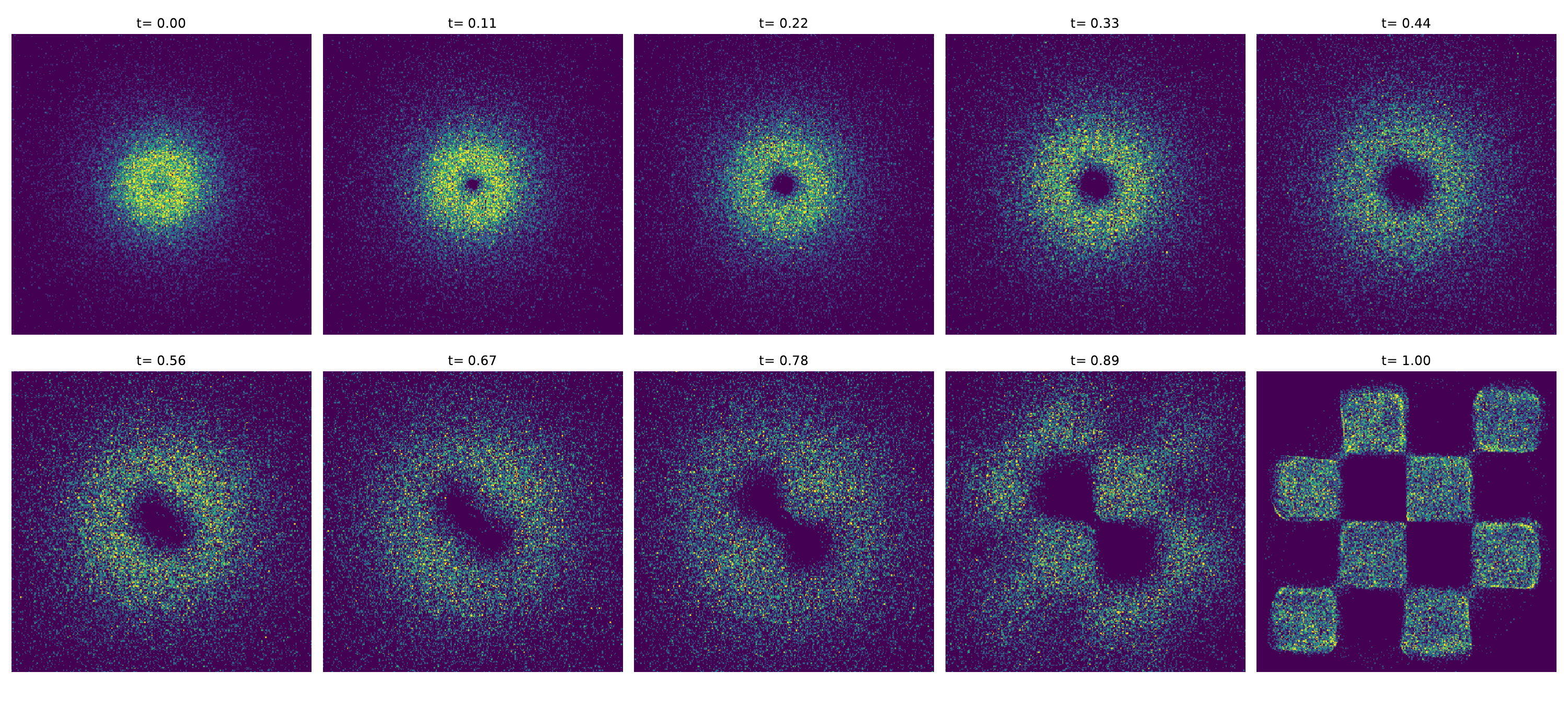"}
    \includegraphics[scale=0.29]{"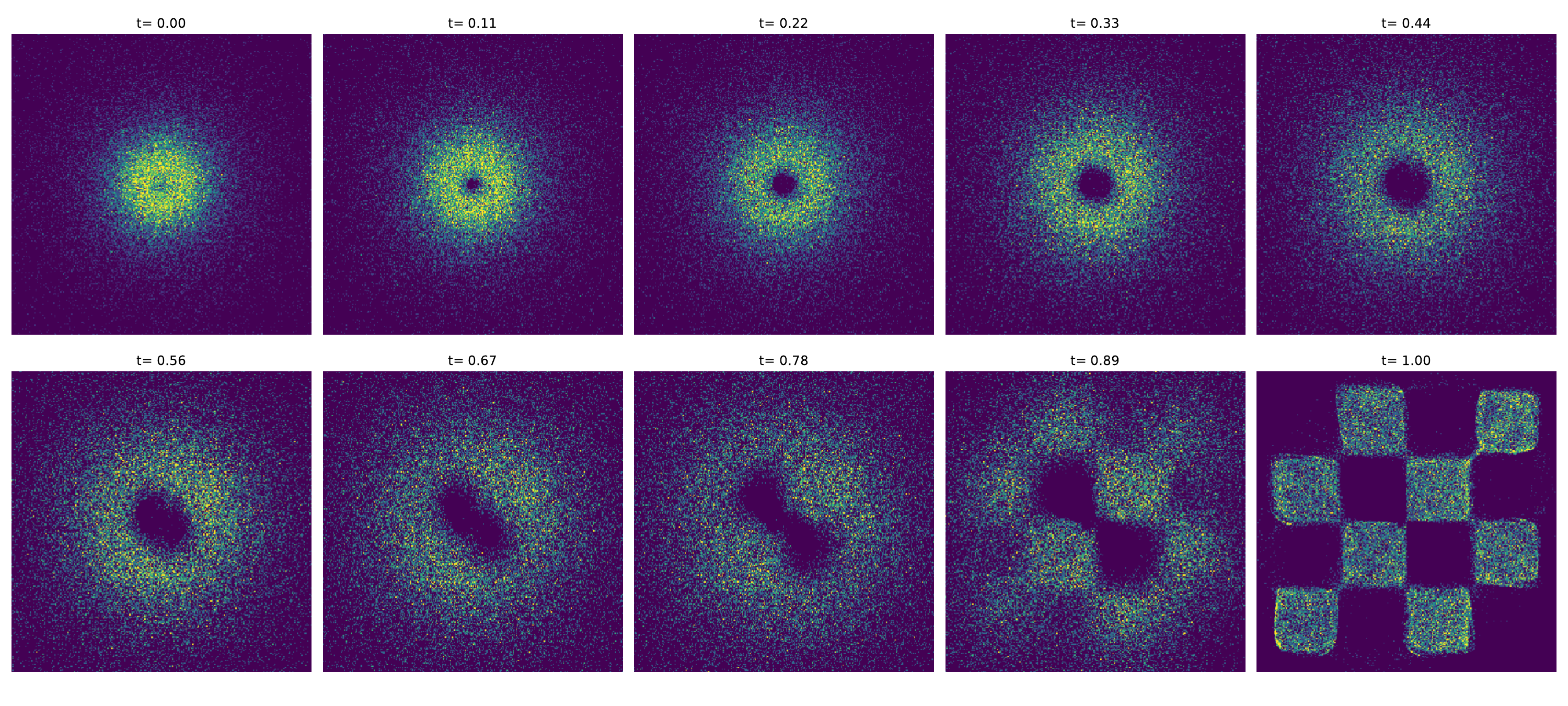"}
    \caption{Flow matching on \( \operatorname{SO}(3, \mathbb{R}) / \operatorname{SO}(2, \mathbb{R}) \cong S^2 \). Top: Lie-agnostic model. Middle: Lie group model. Bottom: Lie group model with parameter encoding.}
\end{figure}

Note that, as expected, the Lie-agnostic version struggles to generalize and it is significantly outperformed by the other two models. Also, although adding the parameter encoding does not make a huge difference in our experiments, we expect it to be useful in higher-dimensional settings.

\section{Conclusions and future work}
\label{sec:conclusions and future work}
We proposed a new description of flow matching on Lie groups as flow matching on Lie algebras and described a general framework to extend it to homogeneous spaces. This allows us to avoid dealing with the potentially complicated geometries of homogeneous spaces by working directly on Lie groups. The latter are much simpler spaces to deal with, as flow matching on them reduces to Euclidean flow matching on their Lie algebra whenever the exponential map is surjective or the data distributions are known to lie in the image of the exponential map. We tested our framework on \( \operatorname{SL}(2, \mathbb{R}) / \operatorname{SO}(2, \mathbb{R}) \cong \mathbb{H} \) and \( \operatorname{SO}(3, \mathbb{R}) / \operatorname{SO}(2, \mathbb{R}) \cong S^2 \) and showed competitive performance.

There remain many directions to be explored. For instance, suppose we are given a dataset consisting of representatives of elements in a homogeneous space \( G/H \). Since we are ultimately interested in distributions on the quotient space, building \( H \)-invariance into the model (see for instance \cite{klein2023}) may prove beneficial. Moreover, the dataset could be very sparse in \( G \), in the sense that each fiber of \( G \to G/H \) could contain very few data points, potentially making training and generalization more challenging. Finally, we have not conducted any experiments on high-dimensional data.

We leave these questions for future work and we hope that our discussion provides valuable insights and a foundation for larger-scale studies.

\bibliographystyle{alpha}
\bibliography{bib}


\end{document}